\documentclass[runningheads]{llncs}

\usepackage{eccv}

\usepackage{eccvabbrv}

\usepackage{graphicx}
\usepackage{booktabs}
\usepackage{svg}
\usepackage{bbding}

\usepackage{bbm}
\usepackage{amssymb}
\usepackage{amsmath}

\usepackage[accsupp]{axessibility}  

\usepackage{hyperref}

\usepackage{orcidlink}

\begin{document}

\title{Eliminating Warping Shakes for \\Unsupervised Online Video Stitching} 

\titlerunning{StabStitch}




\author{Lang Nie\inst{1,2}\orcidlink{0000-0002-7776-889X} \and
Chunyu Lin\inst{1,2}\orcidlink{0000-0003-2847-0349}\thanks{Corresponding author: cylin@bjtu.edu.cn} \and Kang Liao\inst{3}\orcidlink{0000-0001-9429-1096} \and Yun Zhang\inst{4}\orcidlink{0000-0003-4174-886X} \and \\ Shuaicheng Liu\inst{5}\orcidlink{0000-0002-4170-4552} \and  Rui Ai\inst{6} \orcidlink{0000-0002-3224-008X} \and Yao Zhao\inst{1,2}\orcidlink{0000-0002-8581-9554}}

\authorrunning{L. Nie \textit{et\ al.}}

\institute{Institute of Information Science, Beijing Jiaotong University \and
Beijing Key Laboratory of Advanced Information Science and Network \and
Nanyang Technological University
\and Communication University of Zhejiang 
\and University of Electronic Science and Technology of China \and HAOMO.AI
}

\maketitle

\begin{abstract}
  In this paper, we retarget video stitching to an emerging issue, named \textit{\textbf{warping shake}}, when extending image stitching to video stitching. It unveils the temporal instability of warped content in non-overlapping regions, despite image stitching having endeavored to preserve the natural structures. Therefore, in most cases, even if the input videos to be stitched are stable, the stitched video will inevitably cause undesired warping shakes and affect the visual experience. To eliminate the shakes, we propose \textit{\textbf{StabStitch}} to simultaneously realize video stitching and video stabilization in a unified unsupervised learning framework. 
Starting from the camera paths in video stabilization, we first derive the expression of stitching trajectories in video stitching by elaborately integrating spatial and temporal warps. Then a warp smoothing model is presented to optimize them with a comprehensive consideration regarding content alignment, trajectory smoothness, spatial consistency, and online collaboration. 
To establish an evaluation benchmark and train the learning framework, we build a video stitching dataset with a rich diversity in camera motions and scenes. Compared with existing stitching solutions, \textit{StabStitch} exhibits significant superiority in scene robustness and inference speed in addition to stitching and stabilization performance, contributing to a robust and real-time online video stitching system. The codes and dataset are available at \url{https://github.com/nie-lang/StabStitch}.
\end{abstract}

\begin{figure}[t]
	\centering
	\includegraphics[width=0.99\linewidth]{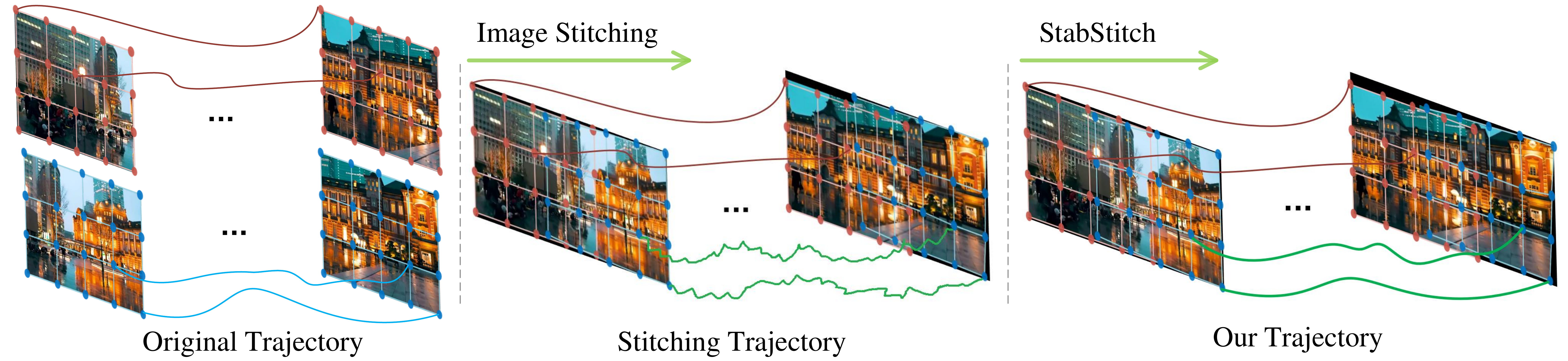}
    \vspace{-0.25cm}
	\caption{The occurrence and elimination of warping shakes. Left: stable camera trajectories for input videos. Middle: warping shakes are produced by image stitching, yielding unsmooth stitching trajectories. Right: \textit{StabStitch} eliminates these shakes successfully.}
	\label{fig1}
 \vspace{-0.3cm}
\end{figure}

\vspace{-0.5cm}
\section{Introduction}
\label{sec:intro}

Video stitching techniques are commonly employed to create panoramic or wide field-of-view (FoV) displays from different viewpoints with limited FoV. Due to their practicality, they are widely applied in autonomous driving \cite{lai2019video}, video surveillance \cite{liu2018future}, virtual reality \cite{lo2021efficient}, etc.
Our work lies in the most common and challenging case of video stitching with hand-held cameras. It does not require camera poses, motion trajectories, or temporal synchronization. It merges multiple videos, whether from multiple cameras or a single camera capturing multiple videos, to create a more immersive representation of the captured scene. Moreover, it transforms video production into an enjoyable and collaborative endeavor among a group of individuals.

Compared with video stitching, image stitching has been studied more extensively and profoundly, which inevitably throws the question of whether existing image stitching solutions can be directly extended to video stitching. Pursuing this line of thought, we initially leverage existing image stitching algorithms \cite{jia2021leveraging}\cite{nie2023parallax} to process hand-held camera videos. Subsequently, we observe that although the stitched results for individual frames are remarkably natural, there is obvious content jitter in the non-overlapping regions between temporally consecutive frames. It is also important to note that the jitter does not originate from the inherent characteristics of the source video itself, although these videos are captured by hand-held cameras. In fact, due to the advancements and widespread adoption of video stabilization in both hardware and software nowadays, the source videos obtained from hand-held cameras are typically stable unless deliberately subjected to shaking. For clarity, we define such content jitter as \textit{warping shake}, which describes the temporal instability of non-overlapping regions induced by temporally non-smooth warps, irrespective of the stability of source videos. Fig. \ref{fig1} illustrates the occurrence process of warping shakes.

Existing video stitching solutions \cite{nie2017dynamic}\cite{su2016video}\cite{guo2016joint}\cite{lin2016seamless}\cite{jiang2015video} follow a strong assumption that each source video from freely moving hand-held cameras suffers from heavy and independent shakes. Consequently, every source video necessitates stabilization via warping, contradicting the current prevalent reality that video stabilization technology has already been widely integrated into various portable devices (\textit{e.g.}, cellphones, DV cameras, and UAVs). In addition, these approaches, to jointly optimize video stabilization and stitching, often establish a sophisticated non-linear solving system consisting of various energy terms. To find the optimal parameters, an iterative solving strategy is typically employed. Each iteration involves several steps dedicated to optimizing different parameters separately, resulting in a rather slow inference speed. Moreover, the complicated optimization procedures also impose stringent requirements on the input video quality (\textit{e.g.}, sufficient, accurate, and evenly distributed matching points), making the video stitching systems fragile and not robust in practical applications.

To solve the above issues, we present the first unsupervised online video stitching framework (termed \textit{StabStitch}) to realize video stitching and video stabilization simultaneously. Building upon the current condition that source videos are typically stable, we simplify this task to stabilize the warped videos by removing warping shakes as illustrated in Fig. \ref{fig1} (right). To get stable stitching warps, we generate the stitching trajectories drawing on the experience of camera trajectories  (\textit{i.e.}, Meshflow \cite{liu2016meshflow}) in video stabilization. By ingeniously combining spatial and temporal warps, we derive the formulation of stitching trajectories in the warped video. Next, a warp smoothing model is presented to simultaneously ensure content alignment, smooth stitching trajectories, preserve spatial consistency, and boost online collaboration. Diverging from conventional offline video stitching approaches that require complete videos as input, \textit{StabStitch} stitches and stabilizes videos with backward frames alone. Besides, its efficient designs further contribute to a real-time online video stitching system with only one frame latency.

As there is no proper dataset readily available, we build a holistic video stitching dataset to train the proposed framework. Moreover, it could serve as a comprehensive benchmark with a rich diversity in camera motions and scenes to evaluate image/video stitching methods. Finally, we summarize our principle contributions as follows:
\begin{itemize}
	\item We retarget video stitching to an emerging issue, termed \textit{warping shake}, and reveal its occurrence when extending image stitching to video stitching.  
	\item We propose \textit{StabStitch}, the first unsupervised online video stitching solution, with a pioneering step to integrating video stitching and stabilization in a unified learning framework.   
    \item We propose a holistic video stitching dataset with diverse scenes and camera motions. The dataset can work as a benchmark dataset and promote other related research work.  
    \item Compared with state-of-the-art image/video stitching solutions, our method achieves comparable or significantly superior performance in terms of scene robustness, inference speed, and stitching/stabilization effect.  
\end{itemize}

\section{Related Work}
Here, we briefly review image stitching, video stabilization, and video stitching techniques, respectively.
 \vspace{-0.2cm}
\subsection{Image Stitching}
Traditional image stitching methods usually detect keypoints \cite{lowe2004distinctive} or line segments \cite{von2008lsd} and then minimize the projective errors to estimate a parameterized warp by aligning these geometric features. To eliminate the parallax misalignment \cite{zhang2014parallax}, the warp model is extended from global homography transformation \cite{brown2007automatic} to other elastic representations, such as mesh \cite{zaragoza2013projective}, TPS \cite{li2017parallax}, superpixel \cite{lee2020warping}, and triangular facet\cite{li2019local}. Meanwhile, to keep the natural structure of non-overlapping regions, a series of shape-preserving constraints is formulated with the alignment objective. For instance, SPHP \cite{chang2014shape} and ANAP \cite{lin2015adaptive} linearized the homography and slowly changed it to the global similarity to reduce projective distortions; DFW \cite{li2015dual}, SPW \cite{liao2019single}, and LPC \cite{jia2021leveraging} leveraged line-related consistency to preserve geometric structures; GSP \cite{chen2016natural} and GES-GSP \cite{du2022geometric} added a global similarity before stitching multiple images together so that the warp of each image resembles a similar transformation as a whole; etc. Besides, Zhang \textit{et\ al.} \cite{zhang2020content} re-formulated image stitching with regular boundaries by simultaneously optimizing alignment and rectangling \cite{he2013rectangling}\cite{nie2022deep}.

Recently, learning-based image stitching solutions emerged. They feed the entire images into the neural network, encouraging the network to directly predict the corresponding parameterized warp model (\textit{e.g.}, homography \cite{nie2020view}\cite{nie2022learning}\cite{jiang2022towards}, multi-homography \cite{song2021end}, TPS \cite{nie2023parallax}\cite{kim2024learning}\cite{zhang2024recstitchnet}, and optical flow \cite{kweon2023pixel}\cite{jia2023learning}). Compared with traditional methods based on sparse geometric features, these learning-based solutions train the network parameters to adaptively capture semantic features by establishing dense pixel-wise optimization objectives. They show better robustness in various cases, especially in the challenging cases where traditional geometric features are few to detect.

\vspace{-0.15cm}
\subsection{Video Stabilization}
Traditional video stabilization can be categorized into 3D \cite{Liu2009content}\cite{liu2012video}, 2.5D \cite{Liu2011subspace}\cite{Goldstein2012video}, and 2D \cite{Matsushita2006full}\cite{Grundmann2011auto}\cite{ma2019effective} methods, according to different motion models. The 3D solutions model the camera motions in 3D space or require extra scene structure for stabilization. The structure is either calculated by structure-from-motion (SfM) \cite{Liu2009content} or acquired from additional hardware, such as a depth camera \cite{liu2012video}, a gyroscope sensor \cite{karpenko2011digital}, or a lightfield camera \cite{smith2009light}. Given the intensive computational demands of these 3D solutions, 2.5D approaches relax the full 3D requirement to partial 3D information. To this end, some additional 3D constraints are established, such as subspace projection \cite{Liu2011subspace} and epipolar geometry \cite{Goldstein2012video}. Compared with them, the 2D methods are more efficient with a series of 2D linear transformations (\textit{e.g.}, affine, homography) as camera motions. To deal with large-parallax scenes, spatially varying motion representations are proposed, such as homography mixture \cite{Grundmann2012calibration}, mesh \cite{liu2013bundled}, vertex profile \cite{liu2016meshflow}, optical flow \cite{liu2014steadyflow}, etc. Moreover, some special approaches focus on specific input (\textit{e.g.}, selfie \cite{yu2018selfie}\cite{yu2021real}, 360 \cite{kopf2016360}\cite{tang2019joint}, and hyperlapse \cite{joshi2015real} videos).

In contrast, learning-based video stabilization methods directly regress unstable-to-stable transformation from data. Most of them are trained with stable and unstable video pairs acquired by special hardware in a supervised manner \cite{wang2018deep}\cite{xu2018deep}\cite{zhang2023minimum}. To relieve data dependence, DIFRINT \cite{choi2020deep} proposed the first unsupervised solution via neighboring frame interpolation. To get a stable interpolated frame, only stable videos are used to train the network. Different from it, DUT \cite{xu2022dut} established unsupervised constraints for motion estimation and trajectory smoothing, learning video stabilization by watching unstable videos. 

\vspace{-0.15cm}
\subsection{Video Stitching}
Video stitching has received much less attention than image stitching. Early works \cite{jiang2015video}\cite{perazzi2015panoramic} stitched multiple videos frame-by-frame, and focused on the temporal consistency of stitched frames. But the input videos were captured by cameras fixed on rigs. For hand-held cameras with free and independent motions, there is a significant increase in temporal shakes. To deal with it, videos were first stitched and then stabilized in \cite{hamza2015stabilization}, while \cite{lin2016seamless} did it in an opposite way (\textit{e.g.}, videos were firstly stabilized, and then stitched). Both of them accomplished stitching or stabilization in a separate step. Later, a joint optimization strategy was commonly adopted in \cite{su2016video}\cite{guo2016joint}\cite{nie2017dynamic}, where \cite{nie2017dynamic} further considered the dynamic foreground by background identification. However, solving such a joint optimization problem regarding stitching and stabilization is fragile and computationally expensive. To this end, we rethink the video stitching problem from the perspective of warping shake and propose the first (to our knowledge) unsupervised online solution for hand-held cameras.

\section{StabStitch}
\label{sec_stabstitch}
\begin{figure}[t]
	\centering
	\includegraphics[width=0.99\linewidth]{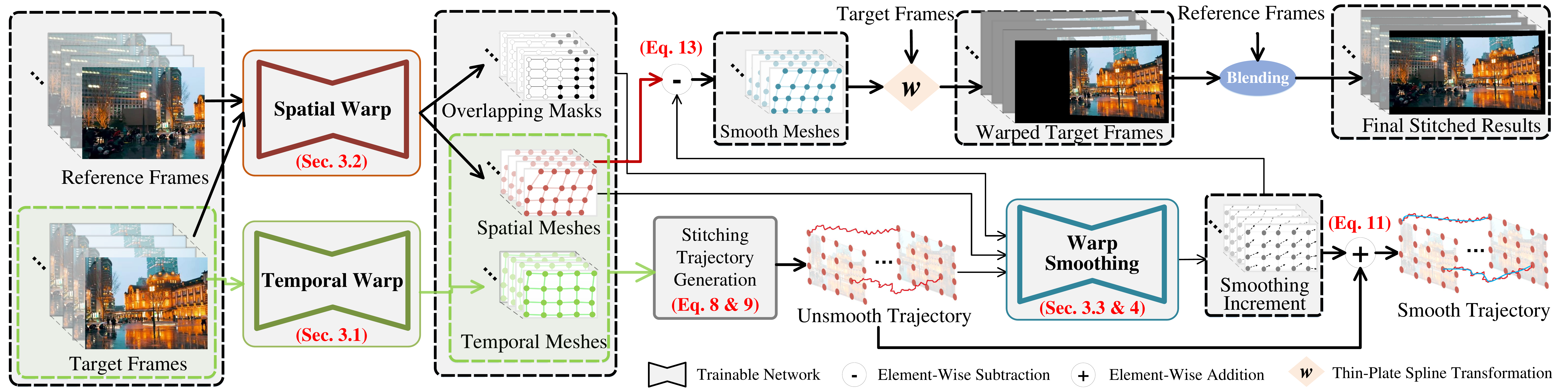}
 \vspace{-0.25cm}
	\caption{The overview of \textit{StabStitch}. We first obtain stitching trajectories by integrating spatial and temporal warps. Then the stitching trajectories are optimized by the warp smoothing model to produce unsmooth-to-smooth stitching warps.}
	\label{network}
  \vspace{-0.3cm}
\end{figure}

We first describe the camera trajectories in video stabilization and then further derive the expression of stitching trajectories in video stitching. Afterward, the unsmooth trajectories are optimized to realize both stitching and stabilization. The pipeline of \textit{StabStitch} is exhibited in Fig. \ref{network}. 

\subsection{Camera Trajectory}
\subsubsection{Temporal Warp:} 
To obtain camera paths, a temporal warp model is first proposed to represent the temporal motion between consecutive video frames. Different from most video stabilization works \cite{xu2022dut}\cite{liu2013bundled}\cite{liu2016meshflow} that use point correspondences to estimate the warp, we leverage a convolutional neural network to capture the high-level information in inter-frame motions. This alternative proves to be robust across various scenarios, particularly in low-light and low-texture environments. The network structure is similar to the warp network of UDIS++ \cite{nie2023parallax}. As shown in Fig. \ref{network}(left), it takes two consecutive target frames as input and outputs the motions of mesh-like control points 
$m(t)$ \cite{bookstein1989principal}. Due to the temporal continuity of adjacent frames in the video, the estimated motions are often not significant. Consequently, we replace all global correlation layers \cite{9605632} in UDIS++ with local correlation layers (\textit{i.e.}, cost volume \cite{sun2018pwc}). To improve the efficiency, we substitute the ResNet50 backbone \cite{he2016deep} in UDIS++ with ResNet18 and reduce network parameters accordingly. Following UDIS++, our optimization objective also consists of an alignment term and a distortion term, as described in the following equation:
\begin{equation}
    \mathcal{L}^{tmp} = \mathcal{L}_{alignment} + \lambda^{tmp} \mathcal{L}_{distortion}.
    \label{eq_temporal_warp}
 \end{equation}
The alignment component leverages photometric errors to implicitly supervise control point motions. The distortion component is composed of an inter-grid constraint and an intra-grid constraint. For brevity, we refer the readers to the supplementary for more details.


\subsubsection{Meshflow:} The camera paths can be defined as a chain of relative motions, such as Euclidean transformations \cite{liu2012video}, homography transformations \cite{liu2013bundled}, etc. Representing the transformation of the initial frame as an identity matrix $F(1)$, the camera trajectories are written as:
\begin{equation}
    C(t)=F(1)F(2)\cdot \cdot \cdot F(t),
    \label{eq_paths_homo}
 \end{equation}
where $F(t)$ is the relative transformation from the $t$-th frame to the $(t-1)$-th frame. 
Considering that our temporal warp model directly predicts the 2D motions of each control point, we adopt the motion representation of vertex files like MeshFlow \cite{liu2016meshflow}. Particularly, we chain the motions of each control point $i$ temporally as the control point trajectory for a more straightforward representation:
\begin{equation}
    C_i(t)=m_i(1)+m_i(2)+\cdot \cdot \cdot +m_i(t),
    \label{eq_paths_meshflow}
 \end{equation}
where $m_i(1)$ is set to zero. Note each control point in $m(t)$ is anchored at every vertex in a rigid mesh. 
 
\subsection{Stitching Trajectory}
\label{sec_Stitching}
Compared with video stabilization, video stitching is more challenging with two or more videos as input and requires the stitched video to possess coherently smooth camera trajectories for the contents from different videos.

\subsubsection{Spatial Warp:} 
To obtain the stitching trajectories, in addition to the temporal warp model, we also establish a spatial warp model to represent the spatial motion between different video views, as shown in Fig. \ref{network}(left). The spatial warp model has a similar network structure to the temporal warp model except that the first local correlation layer is replaced by a global correlation layer \cite{9605632} to capture long-range matching (usually longer than half of the image width/height).
Considering the significance of spatial warping stability in video stitching, we expect this warp to be as robust as possible, although this network model has been proven to be more robust than traditional methods. To this end, we further introduce a motion consistency term in addition to the basic optimization components of the temporal warp:
\begin{equation}
    \mathcal{L}_{consis.}= \frac{_1}{_{(U+1)\times (V+1)}}\sum_{i=1}^{(U+1)\times (V+1)} \Vert m_i(t)-m_i(t-1)-\mu^{spt}\Vert_2,
    \label{eq_motion_consis}
\end{equation}
where $\mu^{spt}$ is the maximum tolerant motion difference and $(U+1)\times (V+1)$ denotes the number of control points. We further sum up the total optimization goal as:
\begin{equation}
    \mathcal{L}^{spt} = \mathcal{L}_{alignment} + \lambda^{spt} \mathcal{L}_{distortion} + \omega^{spt}\mathcal{L}_{consis.}.
    \label{eq_spatial_warp}
 \end{equation}
Refer to the ablation studies or supplementary for the impact of $\mathcal{L}_{consis.}$.

\subsubsection{Stitch-Meshflow:} 
Video stabilization leverages the chain of temporary motions as camera paths, whereas in our video stitching system, how should we represent the stitching paths of a warped video? We dig into this problem by combining the spatial and temporal warp models. With these two models, we first reach the spatial/temporal motions ($m^{S}/m^{T}\in \mathbb{R}^{2\times (U+1) \times (V+1)}$) and their corresponding meshes ($M^{S}/M^{T}\in \mathbb{R}^{2\times (U+1) \times (V+1)}$) as follows:
\begin{equation}
  \label{eq:stitchmeshflow1}
  \begin{matrix}
    \begin{aligned}
      m^{T}&(t) =  TNet(I_{tgt}^{t-1}, I_{tgt}^{t}) \quad\quad \Rightarrow M^{T}(t) = M^{Rig}+m^{T}(t),\\
      m^{S}&(t-1) = SNet(I_{ref}^{t-1}, I_{tgt}^{t-1}) \Rightarrow M^{S}(t-1) = M^{Rig}+m^{S}(t-1),\\
      m^{S}&(t) = SNet(I_{ref}^{t}, I_{tgt}^{t}) \quad\quad\, \,\Rightarrow M^{S}(t) = M^{Rig}+m^{S}(t),
    \end{aligned}\\ 
  \end{matrix}
\end{equation}
where $I_{ref}/I_{tgt}\in \mathbb{R}^{C\times H \times W}$ is the reference/target frame, $SNet$/$TNet(\cdot,\cdot)$ represents the spatial/temporal warp model, and $M^{Rig}\in \mathbb{R}^{2\times (U+1) \times (V+1)}$ is defined as the 2D positions of control points in a rigid mesh. 

Then we need to derive the stitching motion of the warped video from the spatial/temporal meshes. To align the $t$-th frame with the $(t-1)$-th frame in the warped video, the temporal mesh from the $t$-th frame to the $(t-1)$-th frame in the source video ($M^{T}(t)$) should also undergo the same transformation as the spatial warp of the $(t-1)$-th frame ($M^{S}(t-1)$). Assuming $\mathcal{T}(\cdot)$ is the thin-plate spline (TPS) transformation, the desired stitching motion could be represented as the difference between the desired mesh and the actual spatial mesh ($M^{S}(t)$): 
\begin{equation}
    s(t)=\mathcal{T}_{M^{Rig}\to M^{S}(t-1)}(M^{T}(t)) - M^{S}(t).
    \label{eq:stitchmeshflow2}
 \end{equation}
Finally, we attain the stitching paths (we also call it Stitch-Meshflow) by chaining the relative stitching motions between consecutive warped frames as follows:
 \begin{equation}
    S_i(t)=s_i(1)+s_i(2)+\cdot \cdot \cdot +s_i(t),
    \label{eq:stitchmeshflow3}
 \end{equation}
where we define $s(1)$ is an all-zero array.

\subsection{Warp Smoothing}
To get a temporally stable warped video, we need to smooth the stitching trajectories as well as preserve their spatial consistency. Besides, we should also try to prevent the degradation of alignment performance in overlapping areas.

\vspace{-0.3cm}
\subsubsection{Achitecure:}
In this stage, a warp smoothing model is designed to achieve the above goals. As depicted in Fig. \ref{network}, it takes sequences of ($N$ frames) stitching paths ($S$), spatial meshes ($M^S$), and overlapping masks ($OP$) as input, and outputs a smoothing increment ($\Delta$) as described in the following equation:
 \begin{equation}
    \Delta = SmoothNet(S, M^S, OP),
    \label{eq_smoothnet1}
 \end{equation}
where $S/M^S/OP\in \mathbb{R}^{2\times N \times (U+1) \times (V+1)}$. $OP$ are binary mask sequences (1/0 indicates the vertex inside/outside overlapping regions). We calculate it by determining whether each control point in $M^S$ exceeds image boundaries.

The smoothing model first embeds $S$, $M^S$, and $OP$ into 32, 24, and 8 channels through separate linear projections, respectively. Then these embeddings are concatenated and fed into three 3D convolutional layers to model the spatiotemporal dependencies. Finally, we reproject the hidden results back into 2 channels to get $\Delta$. The network architecture is designed rather simply to accomplish efficient smoothing inference. In addition, this simple architecture better highlights the effectiveness of the proposed unsupervised learning scheme.

With the smoothing increment $\Delta$, we define the smooth stitching paths as:
 \begin{equation}
    \hat{S} = S + \Delta.
    \label{eq_smoothnet2}
 \end{equation}
Furthermore, if we expand Eq. \ref{eq_smoothnet2} based on Eq. \ref{eq:stitchmeshflow3} and Eq. \ref{eq:stitchmeshflow2}, we obtain:
\begin{equation}
  \label{eq_smoothnet3}
  \begin{matrix}
    \begin{aligned}
      \hat{S}(t) &=  S(t-1) + s(t) + \Delta(t) \\
      &= S(t-1) + \mathcal{T}_{M^{Rig}\to M^{S}(t-1)}(M^{T}(t)) - \underbrace{(M^{S}(t) - \Delta(t))}_{\text{Smooth spatial mesh}}.
    \end{aligned}\\ 
  \end{matrix}
\end{equation}
In this case, the last term in Eq. \ref{eq_smoothnet3} can be regarded as the smooth spatial mesh $\hat{M}^{S}(t)$. Therefore, the sequences of smooth spatial meshes are written as:
\begin{equation}
    \hat{M}^{S} = M^{S} - \Delta.
    \label{eq_smoothnet4}
 \end{equation}

\subsubsection{Objective Function:}
\label{sec_Optimization}



Given original stitching paths ($S$) and smooth stitching paths ($\hat{S}$), smooth spatial meshes ($\hat{M}^{S}$), and overlapping masks ($OP$), we design the unsupervised learning goal as the balance of different optimization components:
\begin{equation}
    \mathcal{L}^{smooth} =  \mathcal{L}_{data} + \lambda^{smooth} \mathcal{L}_{smothness} + \omega^{smooth} \mathcal{L}_{space}.
    \label{eq_smoothgoal1}
\end{equation}

\paragraph{Data Term:} 
The data term encourages the smooth paths to be close to the original paths. This constraint alone does not contribute to stabilization. The stabilizing effect of \textit{StabStitch} is realized in conjunction with the data term and the subsequent smoothness term. To maintain the alignment performance of overlapping regions during the smoothing process as much as possible, we further incorporate the awareness of overlapping regions into the data term as follows:
\begin{equation}
    \mathcal{L}_{data} = \Vert (\hat{S}-S)(\alpha OP + 1)\Vert_2,
    \label{eq_smoothgoal2}
 \end{equation}
where $\alpha$ is a constant to emphasize the degree of alignment.

\paragraph{Smoothness Term:} 
In a smooth path, each motion should not contain sudden large-angle rotations, and the amplitude of translations should be as consistent as possible. To this end, we constrain the trajectory position at a certain moment to be located at the midpoint between its positions in the preceding and succeeding moments, which implicitly satisfies the above two requirements. Hence, we formulate the smoothness term as:
 \begin{equation}
  \label{eq_smoothgoal3}
  \begin{matrix}
    \begin{aligned}
      \mathcal{L}_{smothness} =  \sum_{j=1}^{(N-1)/2} \beta_{j}\Vert \hat{S}(mid+j) + \hat{S}(mid-j) - 2\hat{S}(mid)\Vert_2,
    \end{aligned}\\ 
  \end{matrix}
\end{equation}
where $mid$ is the middle index of $N$ ($N$ is required to be an odd number) and $\beta_{j}$ is a constant between 0 and 1 to impose varying magnitudes of smoothing constraints on trajectories at different temporal intervals.

\vspace{-0.15cm}
\paragraph{Spatial Consistency Term:} 
When there are only data and smoothness constraints, the warping shakes can be already removed. But each trajectory is optimized individually. Actually, our system has $(U+1)\times(V+1)$ control points, which means there are $(U+1)\times(V+1)$ independently optimized trajectories. When these trajectories are changed inconsistently, significant distortions will be produced. To remove the distortions and encourage different paths to share similar changes, we introduce a spatial consistency component as:
 \begin{equation}
  \label{eq_smoothgoal4}
  \begin{matrix}
    \begin{aligned}
      \mathcal{L}_{space} = \frac{1}{N} \sum_{t=1}^{N} \mathcal{L}_{distortion}(\hat{M}^{S}(t)),
    \end{aligned}\\ 
  \end{matrix}
\end{equation}
where $\mathcal{L}_{distortion}(\cdot)$ takes a mesh as input and calculates the distortion loss like the spatial/temporal warp model.



\begin{figure}[t]
	\centering
	\includegraphics[width=0.99\linewidth]{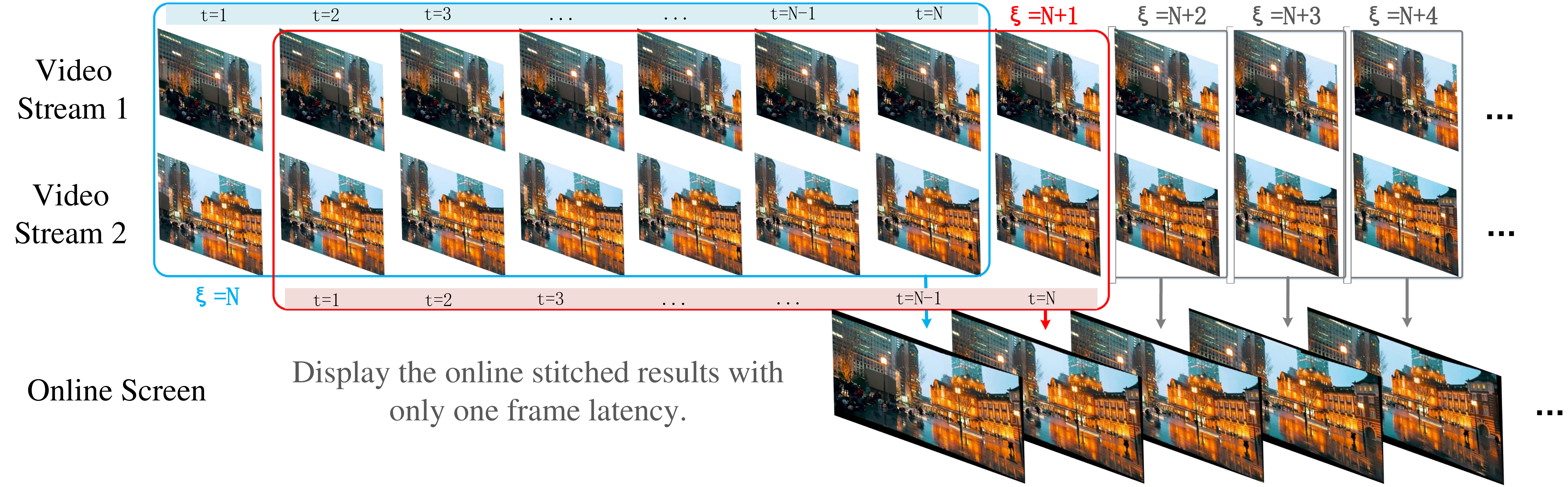}
 \vspace{-0.35cm}
	\caption{The online stitching mode. We define a sliding window to process a short sequence and display the last frame on the online screen.}
	\label{online}
  \vspace{-0.3cm}
\end{figure}

\section{Online Stitching}
\vspace{-0.2cm}
Existing video stitching methods \cite{nie2017dynamic}\cite{su2016video}\cite{guo2016joint}\cite{lin2016seamless}\cite{jiang2015video} are offline solutions, which smooth the trajectories after the videos are completely captured. 
Different from them, \textit{StabStitch} is an online video stitching solution. In our case, the frames after the current frame are no longer available and real-time inference is required.

\vspace{-0.3cm}
\subsection{Online Smoothing}
\vspace{-0.15cm}
To achieve this goal, we define a fixed-length sliding window ($N$ frames) to cover previous and current frames, as shown in Fig. \ref{online}. Then the local stitching trajectory inside this window is extracted and smoothed according to Sec. \ref{sec_stabstitch}. Next, the current target frame is re-synthesized using the optimized smooth spatial mesh (Eq. \ref{eq_smoothnet4}). Finally, we blend it with the current reference frame to get a stable stitched frame and display the result when the next frame arrives. With this mode and efficient architectures, \textit{StabStitch} achieves minimal latency with only one frame.

\paragraph{Online Collaboration Term:} 
However, such an online mode could introduce a new issue, wherein the smoothed trajectories in different sliding windows (with partial overlapping sequences) may be inconsistent. This can produce subtle jitter if we chain the sub-trajectories of different windows. Therefore, we design an online collaboration constraint besides the existing optimization goal (Eq. \ref{eq_smoothgoal1}):
 \begin{equation}
  \label{eq_online1}
  \begin{matrix}
    \begin{aligned}
      \mathcal{L}_{online} = \frac{1}{N-1} \sum_{t=2}^{N} \Vert \hat{S}^{(\xi)}(t) - \hat{S}^{(\xi+1)}(t-1) \Vert_2,
    \end{aligned}\\ 
  \end{matrix}
\end{equation}
where $\xi$ is the absolute time ranging from $N$ to the last frame of the videos. By contrast, $t$ can be regarded as the relative time in a certain sliding window ranging from $1$ to $N$.

\subsection{Offline and Online Inference}
Offline smoothing takes the whole trajectories as input, outputs the optimized whole trajectories, and then renders all the video frames. It carries on smoothing after receiving whole input videos and can be regarded as a special online case in which the sliding window covers whole videos. By contrast, online smoothing takes local trajectories as input, outputs the optimized local trajectories, and then renders the last frame in the sliding window. The online process smoothes current paths without using any future frames.

\begin{figure}[t]
	\centering
	\includegraphics[width=0.99\linewidth]{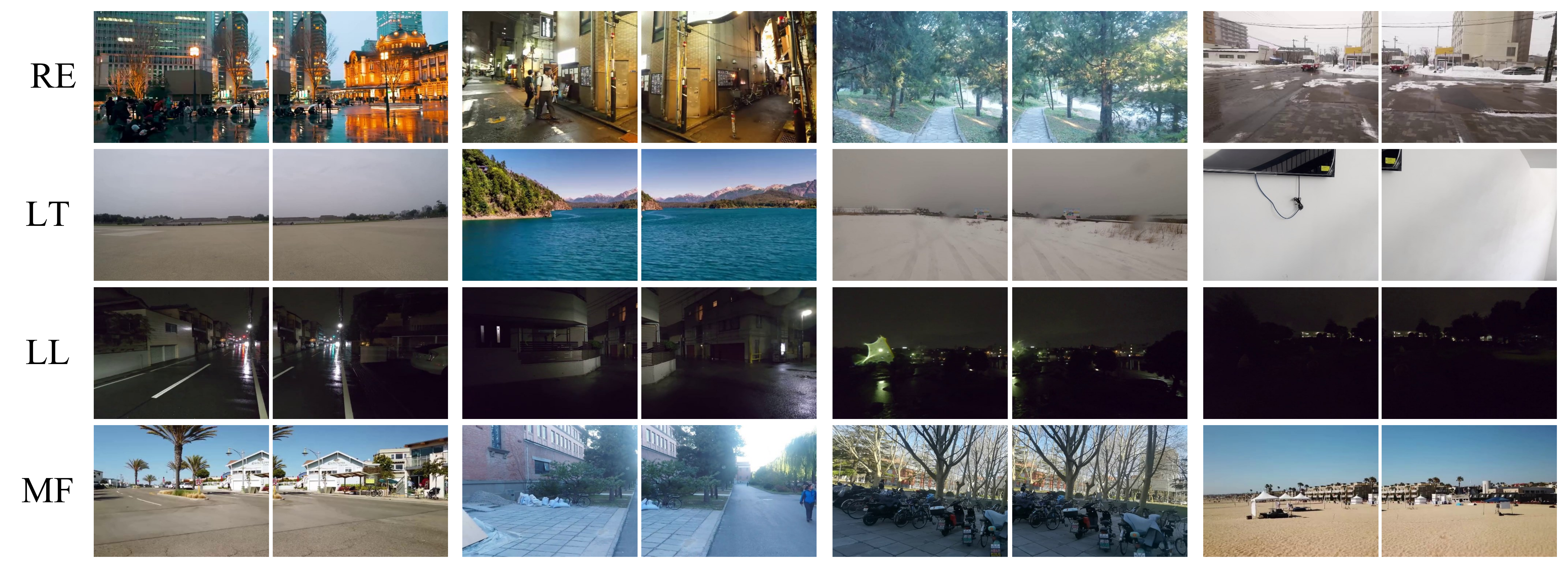}
 \vspace{-0.25cm}
	\caption{The proposed \textit{\textbf{StabStitch-D}} dataset with a large diversity in camera motions and scenes. We exhibit several video pairs for each category.}
	\label{dataset}
  \vspace{-0.3cm}
\end{figure}

\section{Dataset Preparation}
We establish a dataset, named \textit{\textbf{StabStitch-D}}, for the comprehensive video stitching evaluation considering the lack of dedicated datasets for this task. Our dataset comprises over 100 video pairs, consisting of over 100,000 images, with each video lasting from approximately 5 seconds to 35 seconds. To holistically reveal the performance of video stitching methods in various scenarios, we categorize videos into four classes based on their content, including regular (RE), low-texture (LT), low-light (LL), and fast-moving (FM) scenes. In the testing split, 20 video pairs are divided for testing, with 5 videos in each category. Fig. \ref{dataset} illustrates some examples for each category, where FM is the most challenging case with fast irregular camera movements (rotation or translation). The resolution of each video is resized into $360\times 480$ for efficient training, and in the testing phase, arbitrary resolutions are supported.

\section{Experiment}

\subsection{Details and Metrics}
\paragraph{Details:} 
We implement the whole framework in PyTorch with one RTX 4090Ti GPU. The spatial warp, temporal warp, and warp smoothing models are trained separately with the epoch number set to 55, 40, and 50, respectively. $\lambda^{tmp}$, $\lambda^{spt}$, $\mu^{spt}$, and $\omega^{spt}$ are defined as 5, 10, 20, and 0.1. The weights for data, smoothness, spatial consistency, and online collaboration terms are set to 1, 50, 10, and 0.1. $\alpha$, $\beta_1$, $\beta_2$, and $\beta_3$ are set to 10, 0.9, 0.3, and 0.1. The control point resolution and sliding window length are set to $(6+1)\times(8+1)$ and 7. Moreover, when training the warp smoothing model, we randomly select $N=7$ frames as the processing window from a larger buffer of 12 frames, which could allow more diverse stitching paths.

\paragraph{Metrics:} 
To quantitatively evaluate the proposed method, we suggest three metrics including \textbf{alignment score}, \textbf{distortion score}, and \textbf{stability score}. Limited by space, we moved the related metric details to the supplementary.

\subsection{Compared with State-of-The-Arts} 
\label{results}
We compare our method with image and video stitching solutions, respectively.

\vspace{-0.3cm}
\subsubsection{Compared with Image Stitching:}
Two representative SoTA image stitching methods are selected to compare with our solution: LPC \cite{jia2021leveraging} (traditional method) and UDIS++ \cite{nie2023parallax} (learning-based method).
The quantitative comparison results are illustrated in Tab. \ref{quantitative results}, where `$\cdot/\cdot$' indicates the PSNR/SSIM values. `-' implies the approach fails in this category (\textit{e.g.}, program crash and extremely severe distortion). The results show our solution achieves comparable alignment performance with SoTA image stitching methods. In fact, our spatial warp model has surpassed UDIS++ as indicated in Tab. \ref{ablation results}. \textit{StabStitch} sacrifices a little alignment performance to reach better temporally stable sequences.


\setlength{\tabcolsep}{4pt}
\begin{table}[t]
	\begin{center}
		\renewcommand{\arraystretch}{1.1}
		\caption{Quantitative comparisons with image stitching methods on StabStitch-D dataset. * indicates the model is re-trained on the proposed dataset.}
        \vspace{-0.4cm}
		\label{quantitative results}
    \scalebox{0.92}{
		\begin{tabular}{ccccccc}
			\hline
			 &Method & Regular & Low-Light & Low-Texture & Fast-Moving & Average  \\
			\cline{2-7}
            1 & LCP \cite{jia2021leveraging}  & 24.22/0.812 & - & - & 23.88/0.813 & - \\
			2 & UDIS++ \cite{nie2023parallax}  & 23.19/0.785 & 31.09/0.936 & 29.98/0.906 & 21.56/0.756 & 27.19/0.859 \\
            3 & UDIS++ * \cite{nie2023parallax}  & 24.63/0.829 & 34.26/0.957 & 32.81/0.920 & $\mathbf{24.78}$/$\mathbf{0.819}$ & 29.78/$\mathbf{0.891}$ \\
            4 & StabStitch  & $\mathbf{24.64}$/$\mathbf{0.832}$ & $\mathbf{34.51}$/$\mathbf{0.958}$ & $\mathbf{33.63}$/$\mathbf{0.927}$ & 23.36/0.787 & $\mathbf{29.89}$/0.890 \\
			\hline
		\end{tabular}
  }
	\end{center}
 \vspace{-0.5cm}
\end{table}

\begin{table}[!t]

  \centering
  \caption{ User study on the cases that Nie \textit{et\ al.} \cite{nie2017dynamic} successes, in which the user preference rate is reported. We exclude the failure cases of Nie \textit{et\ al.} \cite{nie2017dynamic} for fairness.}
  \vspace{-0.2cm}
  \renewcommand{\arraystretch}{1.2}
  \begin{tabular}{ccc}
   \hline
    StabStitch &  Nie \textit{et\ al.} \cite{nie2017dynamic} & No preference \\
   \hline
  30.47\% & 6.25\%  & 63.28\% \\
      \hline
   \end{tabular}
   \vspace{-0.4cm}
   \label{table:user}
   \end{table}

\vspace{-0.3cm}
\subsubsection{Compared with Video Stitching:} We compare our method with Nie \textit{et\ al.}'s video stitching solution \cite{nie2017dynamic}. To our knowledge, it is the latest and best video stitching method for hand-held cameras. Based on the assumption that the input videos are unstable, it estimates two respective non-linear warps for the reference and target video frame. In contrast, we hold the assumption that currently input videos are typically stable unless deliberately subjected to shaking. Only the target video frame is warped in our system. This difference between Nie \textit{et\ al.} \cite{nie2017dynamic} and our solution makes the comparison of PSNR/SSIM not completely fair. Therefore, we conduct a user study as an alternative and demonstrate extensive stitched videos in our supplementary video.

\begin{figure}[t]
	\centering
    
	\includegraphics[width=0.99\linewidth]{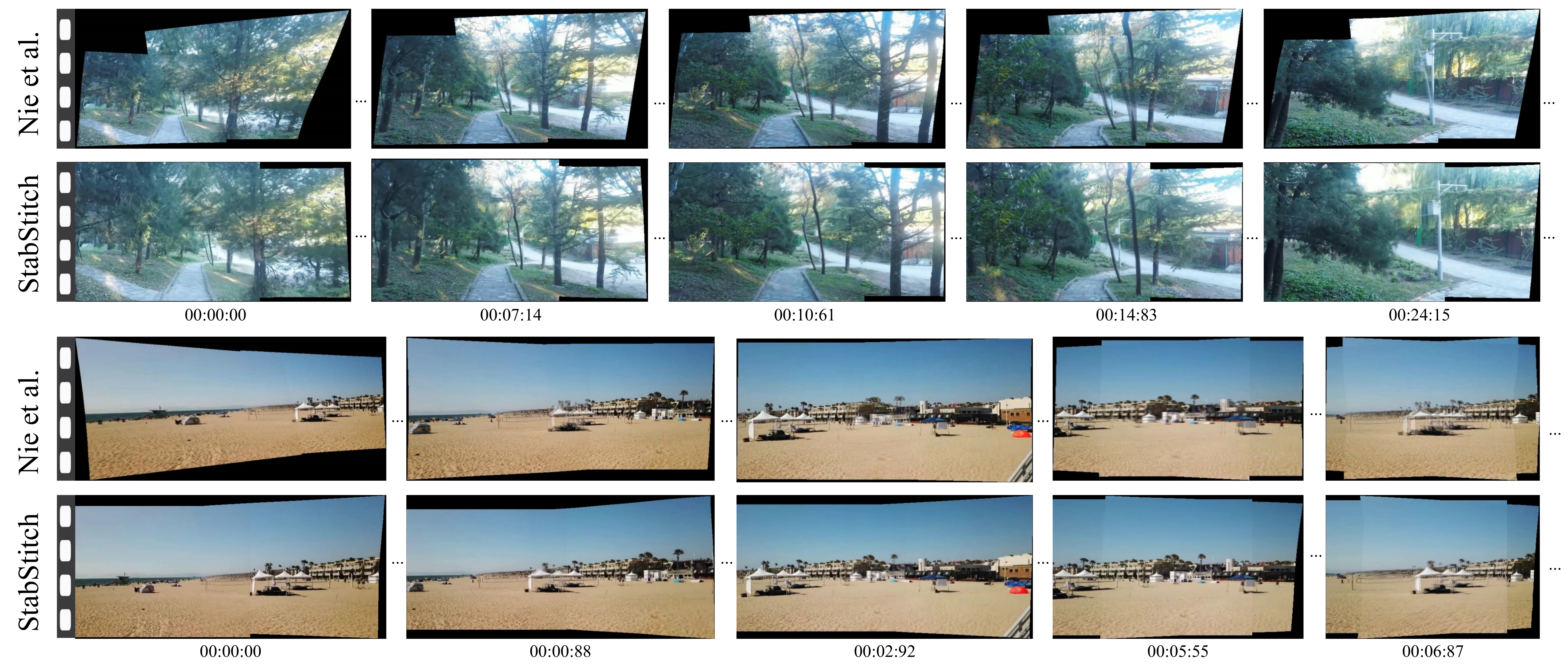}
 \vspace{-0.25cm}
	\caption{Qualitative comparison with Nie \textit{et\ al.}'s video stitching \cite{nie2017dynamic} on a regular case (top) and a fast-moving case (bottom). The numbers below the images indicate the time at which the frame appears in the video. Please zoom in for the best view.}
	\label{fig:comparison}
\end{figure}

\paragraph{User Preference:} Nie \textit{et\ al.}'s solution \cite{nie2017dynamic} is sensitive to different scenes. In our testing set (20 pairs of videos in total), Nie \textit{et\ al.} \cite{nie2017dynamic} fail in 10 pairs of videos because of program crashes (mainly appearing in the categories of LL and LT). Hence, we exclude these failure cases and conduct the user study only on the successful cases. For a stitched video, different methods may perform differently at different times. So, we segment each complete stitched video into one-second clips (we omit the last clip of a stitched video that is shorter than one second in practice), resulting in 128 clips in total. Then we invite 20 participants, including 10 researchers/students with computer vision backgrounds and 10 volunteers outside this community. In each test session, two clips from different methods are presented in a random order, and every volunteer is required to indicate their overall preference for alignment, distortion, and stability. We average the preference rates and exhibit the results in Tab. \ref{table:user}. From that, our results are more preferred by users. Besides, we illustrate two qualitative examples in Fig. \ref{fig:comparison}, where our results show much fewer artifacts (refer to our supplementary video for the complete stitched videos).

\begin{table}[!t]
  \centering
  \caption{A comprehensive analysis of inference speed (/ms).}
  \vspace{-0.2cm}
  \renewcommand{\arraystretch}{1.2}
  \begin{tabular}{ccccccc}
   \hline
    SNet &  TNet & Trajectory generation & SmoothNet & Warping & Blending & Total \\
   \hline
   11.5 &  10 &  1.1 & 1 & 4.4 & 0.2& 28.2\\
      \hline
   \end{tabular}
   \vspace{-0.3cm}
   \label{table:speed}
\end{table}


\paragraph{Inference Speed:} A comprehensive analysis of our inference speed is shown in Tab. \ref{table:speed} with one RTX 4090Ti GPU, where `Blending' represents the average blending. In the example shown in Fig. \ref{fig:comparison}(top), \textit{StabStitch} only takes about 28.2ms to stitch one frame, yielding a real-time online video stitching system. When stitching a video pair with higher resolution, only the time for warping and blending steps will slightly increase. In contrast, Nie \textit{et\ al.}'s solution \cite{nie2017dynamic} takes over 40 minutes to get such a 26-second stitched video with an Intel i7-9750H 2.60GHz CPU, making it impractical to be applied to online stitching.

\subsection{Ablation Study}

\setlength{\tabcolsep}{4pt}
\begin{table}[t]
	\begin{center}
		\renewcommand{\arraystretch}{1.1}
		\caption{Ablation studies on alignment, distortion, and stability.}
        \vspace{-0.2cm}
		\label{ablation results}
    \scalebox{0.92}{
		\begin{tabular}{ccccccc}
			\hline
			 & Basic Stitching & $\mathcal{L}_{consis.}$ & Warp Smoothing & Alignment $\uparrow$ & Distortion $\downarrow$ & Stability $\downarrow$  \\
			\cline{2-7}
            1 & \Checkmark & & & 30.67/0.902 & 0.784 & 81.57  \\
            2 & \Checkmark & \Checkmark &  & $\mathbf{30.75}$/$\mathbf{0.903}$ & 0.804  & 60.32  \\
            3 & \Checkmark & \Checkmark &  \Checkmark & 29.89/0.890 & $\mathbf{0.674}$ & $\mathbf{48.74}$ \\
			\hline
		\end{tabular} 
  }
	\end{center}
  \vspace{-0.5cm}
\end{table}

\begin{figure}[t]
	\centering
	\includegraphics[width=0.99\linewidth]{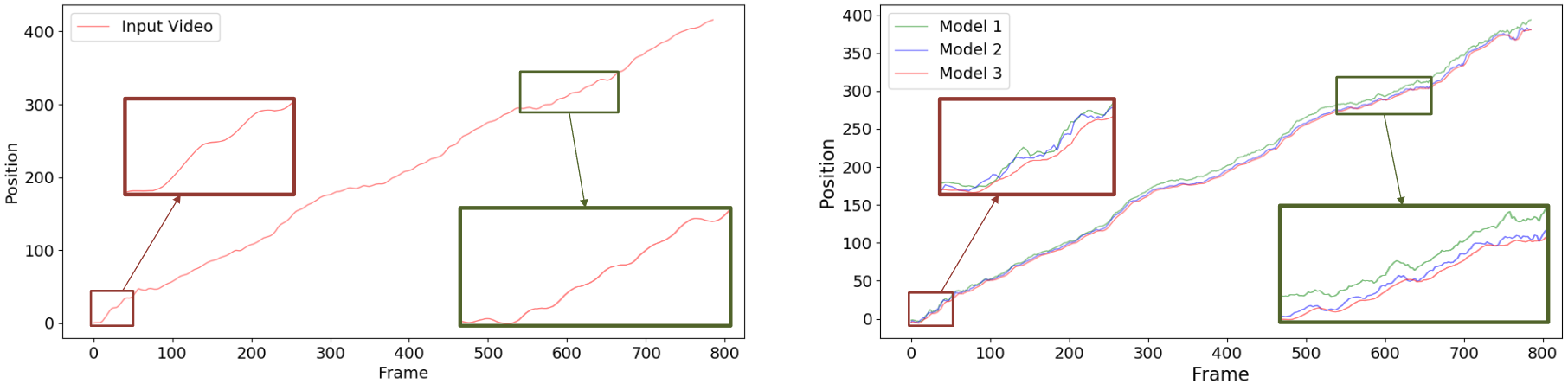}
 \vspace{-0.35cm}
	\caption{Left: camera trajectories of the original target video. Right: stitching trajectories of the warped target video from different models (the model index corresponds to the experiment number in Tab. \ref{ablation results}).}
	\label{trajectory}
  \vspace{-0.3cm}
\end{figure}

\paragraph{Quantitative Analysis:} The main ablation study is shown in Tab. \ref{ablation results}, where `Basic Stitching' (model 1) refers to the spatial warp model without the motion consistency term $\mathcal{L}_{consis.}$. With $\mathcal{L}_{consis.}$ (model 2), the stability is improved. With the warp smoothing model (model 3), both the distortion and stability are significantly optimized at the cost of slight alignment performance, achieving an optimal balance of alignment, distortion, and stability. More experiments can be found in the supplementary materials.

\paragraph{Trajectory Visualization:} We visualize the trajectories of the original target video and warped target videos in Fig. \ref{trajectory}. Here, the trajectories are extracted from a control point of the example shown in Fig. \ref{fig:comparison}(top) in the vertical direction. It can be observed that even if the input video is stable, image stitching can introduce undesired warping shakes, whereas \textit{StabStitch} (Model 3) minimizes these shakes as much as possible during stitching.

\section{Conclusions}
Nowadays, the videos captured from hand-held cameras are typically stable due to the advancements and widespread adoption of video stabilization in both hardware and software. Under such circumstances, we retarget video stitching to an emerging issue, \textit{warping shake}, which describes the undesired content instability in non-overlapping regions especially when image stitching technology is directly applied to videos. To solve this problem, we propose the first unsupervised online video stitching framework, \textit{StabStitch}, by generating stitching trajectories and smoothing them. Besides, a video stitching dataset with various camera motions and scenes is built, which we hope can work as a benchmark and promote other related research work. Finally, we conduct extensive experiments to demonstrate our superiority in stitching, stabilization, robustness, and speed.


\vspace{0.3cm}
\noindent\textbf{Acknowledgments:} This work was supported by the National Natural Science Foundation of China (No. 62172032), and Zhejiang Province Basic Public Welfare Research Program (No. LGG22F020009).


%
%
\bibliographystyle{splncs04}
\bibliography{main}

\clearpage
\appendix

\section{Appendix}
\label{sec:1}
In this document, we provide the following supplementary contents:
\begin{itemize}
    \item Details of the spatial/temporal warp model (Section \ref{sec:2}).
    \item Details of the warp smoothing model (Section \ref{sec:3}).
    \item Details of dataset distribution (Section \ref{sec:444}).
    \item Evaluation metric (Section \ref{sec:5}).
    \item More experiments (Section \ref{sec:6}).
 \end{itemize}

Although we present more network details in this supplementary, we argue that these network architectures themselves are not the primary contribution of this work (although we appropriately modified them and achieved improvements). Our main contribution lies in the new paradigm of unsupervised online video stitching, including the representation of stitching trajectories and the design of unsupervised smoothing optimization objectives.

For clarity, we summarize a part of notations and their corresponding meanings in Table \ref{table:notation}.

Besides, we also provide a supplementary video. Please refer to \url{https://www.youtube.com/watch?v=03kGEZJHxzI} for the stitched videos from different methods.

\section{Spatial/Temporal Warp Model}
\label{sec:2}
Due to the similarity to UDIS++ \cite{nie2023parallax}, we just briefly described the structure and loss function of the spatial/temporal model in our manuscript. Here, we give more details in the supplementary material.

We first review the warp model of UDIS++ \cite{nie2023parallax} in Fig. \ref{fig11}(a) and then depict the differences. UDIS++ \cite{nie2023parallax} adopts ResNet50 as the backbone and predicts the control point motions in two steps. The first step estimates the 4-pt homography motions \cite{detone2016deep} and converts them as the initial control point motions, while the second step estimates the residual control point motions, which could reach the final control point motions by addition with initial motions. Both steps leverage the global correlation layer (\textit{i.e.}, the contextual correlation layer \cite{9605632}) to capture feature matching information and then regress the motions with simple regression networks.

\subsection{Structure Difference}

\begin{figure}[!t]
	\centering
	\includegraphics[width=0.99\linewidth]{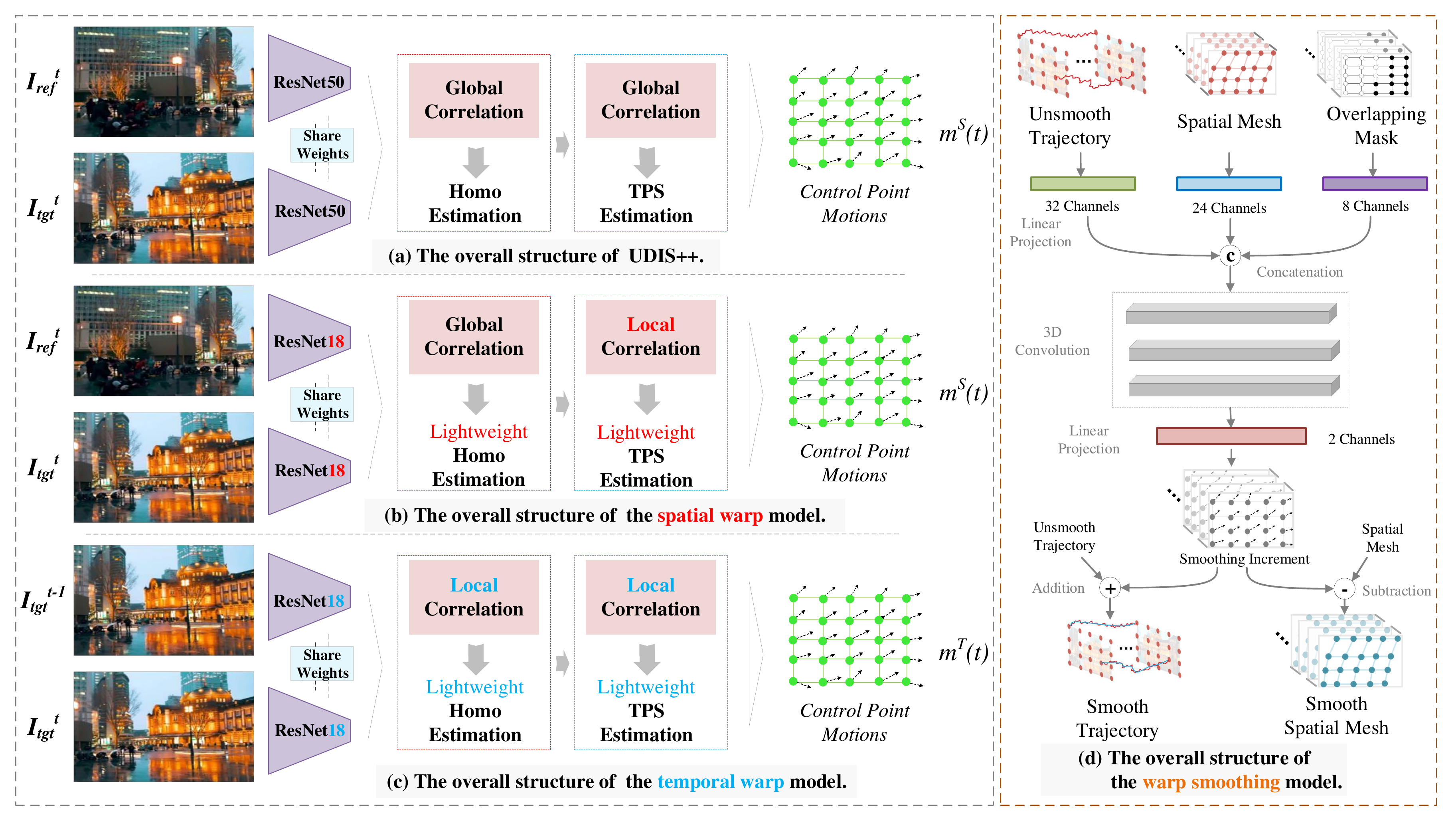}
    \vspace{-0.25cm}
	\caption{The overall structures of our models. Left: the spatial/temporal warp model. Right: the warp smoothing model.}
	\label{fig11}
 \vspace{-0.1cm}
\end{figure}

We demonstrate the structure differences between the spatial/temporal warp model and UDIS++ in Fig. \ref{fig11}(b)/(c). The differences are highlighted in {\color{red}{red}}/{\color{blue}{blue}}. The local correlation layer denotes the cost volume layer \cite{sun2018pwc}. In the spatial warp model, the search radius of the local correlation layer is set to 5, while in the temporal warp model, we set the radius to 6 and 3.

Besides, we further simplify the network architecture, especially the regression networks, significantly reducing the network parameters. For clarity, we compare the model size and report them in Tab. \ref{table:size}.


\begin{table}[!t]

  \centering
  \caption{ The notation table.}
  \vspace{-0.2cm}
  \renewcommand{\arraystretch}{1.2}
  \scalebox{0.8}{
  \begin{tabular}{ccc}
   \hline
    Notation &  Meaning & Example \\
   \hline
  $(t)$ & The relative time in a sliding window.  & \textit{e.g.}, C(t), m(t), S(t)\\
  Subscript $i$ & The control point index.  & \textit{e.g.}, $C_i(t)$, $m_i(t)$, $S_i(t)$\\
  Superscript $T$, $S$ & From the temporal or spatial model.  & \textit{e.g.}, $m^{T}_i(t)$, $M^{S}_i(t)$\\
  Superscript $(\xi)$ & The absolute time ranging from $N$ to the last frame.  & \textit{e.g.}, $m^{T}_i(t)$, $M^{S}_i(t)$\\
      Hat $\hat{}$& The optimized mesh or trajectory.  & \textit{e.g.}, $\hat{M}^{S}(t)$, $\hat{S}$\\
      $M^{Rig}$& The rigid and regular initial mesh (predefined).  & \\
$I_{ref}^{t}/I_{tgt}^{t}$& The $t$-th reference/target frame.  & \\
$SNet/TNet/SmoothNet$& The spatial/temporal/smooth warping models.  & \\
$\mathcal{T}_{M^{Rig}\to M^{S}(t-1)}(\cdot)$& The TPS transformation from $M^{Rig}$ to $M^{S}(t-1)$.  & \\

      \hline
   \end{tabular}
   }
   \vspace{-0.4cm}
   \label{table:notation}
   \end{table}

\begin{table}[!t]
  \centering
  \caption{Model size (/MB).}
  \vspace{-0.2cm}
  \renewcommand{\arraystretch}{1.2}
  \begin{tabular}{ccccc}
   \hline
     &SNet &  TNet  & SmoothNet & Total \\
   \hline
    \textit{StabStitch}& 28.55 &  28.76 &  2.12 & 59.43   \\
    UDIS++ \cite{nie2023parallax} & 297.73 &  -  & - & 297.73 \\
   \hline
   \end{tabular}
   \vspace{-0.3cm}
   \label{table:size}
\end{table}

\subsection{Loss Function}
The alignment loss and distortion loss for the spatial/temporal warp model are also similar to UDIS++ \cite{nie2023parallax}. One can refer to \cite{nie2023parallax}\cite{nie2022deep} for more details. For the convenience of readers, we paraphrase their definitions here again.

\paragraph{Alignment Loss:}
As described above, the spatial/temporal warp model takes two steps to predict the final control point motions $m^S(t)$/$m^T(t)$ from global homography transformation to local TPS transformation. Assuming the estimated warping functions for homography and thin-plate spline are represented as $\mathcal{W}_{H}(\cdot)$ and $\mathcal{W}_{T}(\cdot)$, the alignment loss is written as:
\begin{equation}
\begin{aligned}
    \mathcal{L}_{alignment} = \omega_H \Vert I_{ref}\cdot\mathcal{W}_{H}(\mathbbm{1})- &\mathcal{W}_{H}(I_{tgt})\Vert_1 + 
     \omega_H \Vert I_{tgt}\cdot\mathcal{W}_{H^{-1}}(\mathbbm{1})-\mathcal{W}_{H^{-1}}(I_{ref})\Vert_1 \\+ 
     &\Vert I_{ref}\cdot\mathcal{W}_{T}(\mathbbm{1})- \mathcal{W}_{T}(I_{tgt})\Vert_1, 
    \end{aligned}
\end{equation}
where $I_{ref}$/$I_{tgt}$ is the reference/target frame, $\mathbbm{1}$ is an all-one matrix with the same size as $I_{ref}$, and $\omega_H$ is a constant to balance different transformations.

\begin{figure}[!t]
	\centering
	\includegraphics[width=0.85\linewidth]{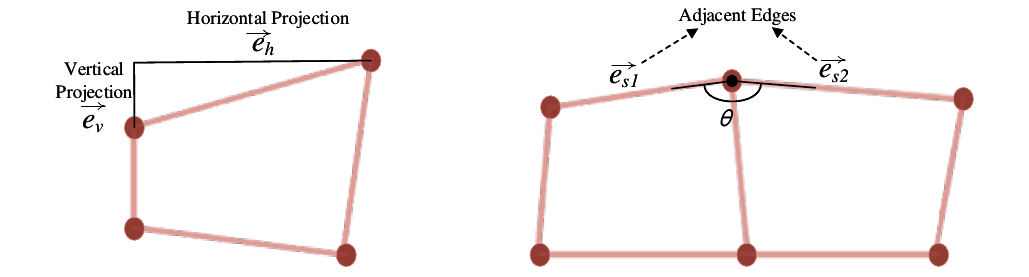}
    \vspace{-0.1cm}
	\caption{The intra-grid (left) and inter-grid (right) constraints in the distortion loss.}
	\label{grid}
 \vspace{-0.3cm}
\end{figure}

\paragraph{Distortion Loss:}
The distortion loss consists of an intra-grid constraint and an inter-grid constraint as follows:
\begin{equation}
\begin{aligned}
    \mathcal{L}_{distortion} = \ell_{intra} + \ell_{inter}. 
\end{aligned}
\end{equation}

The intra-grid term prevents projection distortion caused by excessively large grids after warping by penalizing the grids with side lengths exceeding a certain threshold. As shown in Fig. \ref{grid}(left), we define the horizontal/vertical projection of each grid edge as $\vec{e_h}$/$\vec{e_v}$ and the corresponding projection set as $\{\vec{e_h}\}$/$\{\vec{e}_{v}\}$. Then we can define the intra-grid loss as:
\begin{equation}
\begin{aligned}
  \ell_{intra}= \frac{_1}{_{(U+1)\times V}}\sum_{\{\vec{e_h}\}}\sigma(\vec{e_h}-\frac{_{2W}}{^V}) 
  +  \frac{_1}{_{U\times (V+1)}}\sum_{\{\vec{e}_{v}\}}\sigma(\vec{e_v}-\frac{_{2H}}{^U}),
\end{aligned}
  \label{eq7}
\end{equation}
where $H\times W$ and $(U+1)\times (V+1)$ are the image and control point resolutions. $\sigma(\cdot)$ is the $ReLU$ activation function.

As for the inter-grid term, it is used to reduce structural distortion caused by inconsistent changes in adjacent grid edges (denoted by $\vec{e}_{s1},\vec{e}_{s2}$). As shown in Fig. \ref{grid}(right), if the changes in adjacent edges are consistent, the included angle should be close to $180°$. Therefore, we encourage its cosine distance to approximate $1$ as follows:
\begin{equation}
\begin{aligned}
  \ell_{inter}= \frac{1}{Q}\sum_{\{\vec{e}_{s1}, \vec{e}_{s2}\}}(1-\frac{\langle \vec{e}_{s1},\vec{e}_{s2}\rangle}{\parallel \vec{e}_{s1}\parallel \cdot \parallel \vec{e}_{s2}\parallel }),
  \end{aligned}
\end{equation}
where $\{\vec{e}_{s1}, \vec{e}_{s2}\}$ and $Q$ are the set of horizontal and vertical adjacent edges and their number.

\section{Warp Smoothing Model}
\label{sec:3}
The network structure of the warp smoothing model is exhibited in Fig. \ref{fig11}(d). Although its architecture is very simple (merely consisting of several fully connected layers and 3D convolutions), it can still achieve good results with effective and reasonable loss constraints.


\section{Dataset}
\label{sec:444}
\begin{figure}[!t]
	\centering
	\includegraphics[width=0.7\linewidth]{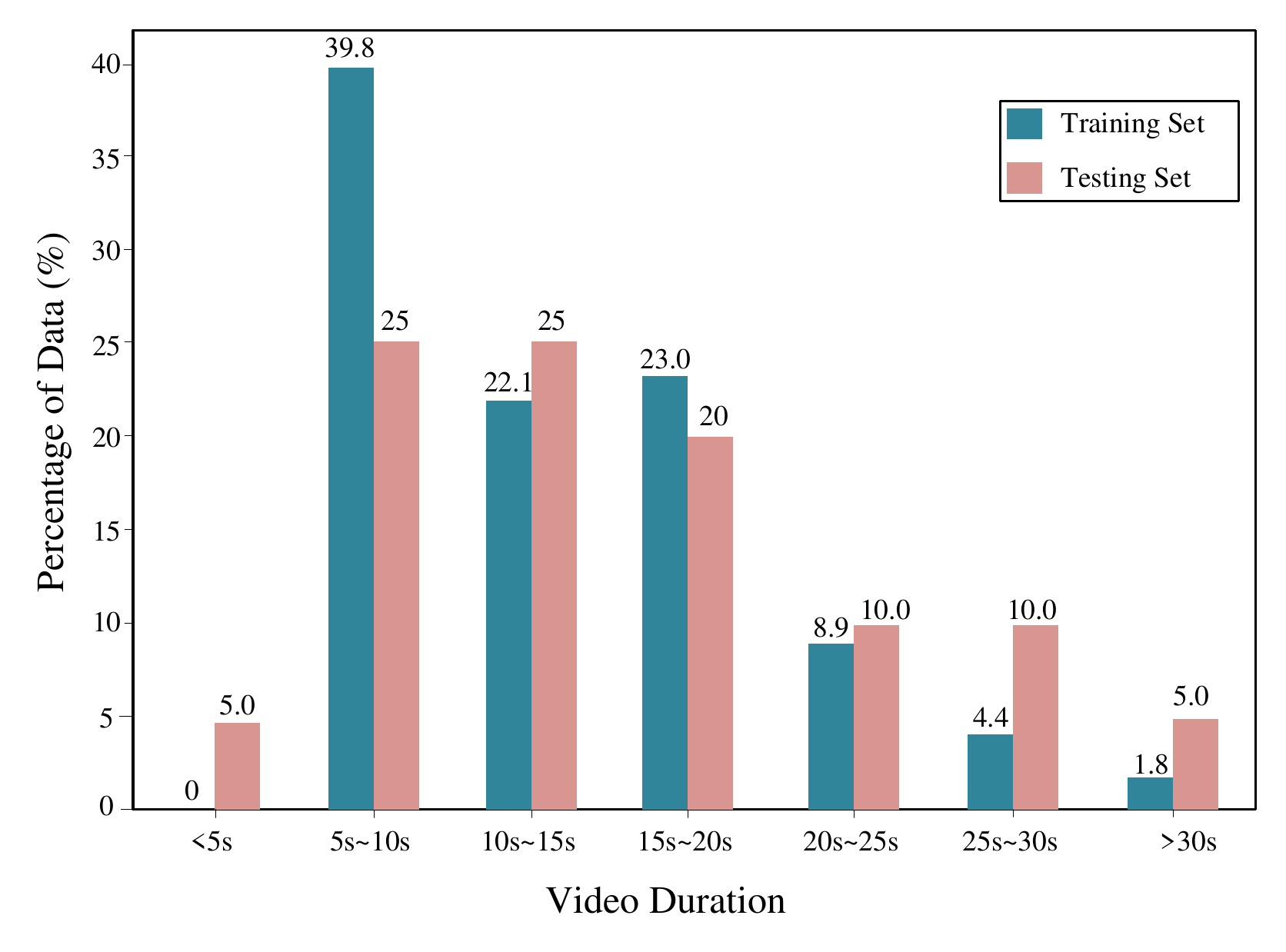}
    \vspace{-0.1cm}
	\caption{The distribution statistics of the video duration time.}
	\label{dataset_duration}
 \vspace{-0.3cm}
\end{figure}

The videos in our dataset consist of three parts: some original videos from \cite{zhang2020content}, some stable videos from \cite{wang2018deep}, and our captured videos. These videos are captured with arbitrary and irregular motion trajectories. Therefore, we leverage them to simulate two videos from different perspectives. Specifically, we collect the video pair from different timestamps (\textit{e.g.}, one video is from the original video, and the other video is captured after a random delay time). After that, we crop the video frames to simulate an appropriate overlapping rate in stitching. Considering the videos are collected from different timestamps, we further filter out the videos with obvious moving objects.
Finally, we get over 100 video pairs and demonstrate the distribution of video duration in Fig. \ref{dataset_duration}.  

\section{Evaluation Metric}
\label{sec:5}
To quantitatively evaluate the proposed method, we suggest three metrics as described in the following:

\noindent\textbf{Alignment Score}: Following the criterion of UDIS \cite{nie2021unsupervised} and UDIS++ \cite{nie2023parallax}, we also adopt PSNR and SSIM of the overlapping regions to evaluate the alignment performance. We average the scores in all video frames.

\noindent\textbf{Distortion Score}: The final warp in the online stitching mode can be described as a series of meshes: $\hat{M}^{S(N)}(N)$, $\hat{M}^{S(N+1)}(N)$, $\cdot\cdot\cdot$, $\hat{M}^{S(\xi)}(N)$, $\cdot\cdot\cdot$. Then we adopt $\mathcal{L}_{distortion}(\cdot)$ to measure the distortion magnitude. Because any distortion in a single frame will destroy the perfection of the whole result, we choose the mean value of the maximum distortion loss of each video as the distortion score. 

\noindent\textbf{Stability Score}: The smoothed trajectories in the online stitching mode can also be described as a series of positions: $\hat{S}^{(N)}(N)$, $\hat{S}^{(N+1)}(N)$, $\cdot\cdot\cdot$, $\hat{S}^{(\xi)}(N)$, $\cdot\cdot\cdot$. Then we adopt $\mathcal{L}_{smoothness}(\cdot)$ to measure the stability. The stability score is the mean value of the average smoothness loss of each video.

Please note that in the comparative experiments, we only adopt the alignment score because different methods define different warp representations and camera trajectories. Thus we only apply the last two metrics to our ablation studies to show the effectiveness of each module.

In the beginning, we evaluate the distortion and stability performance with the metrics that are widely used in video stabilization \cite{liu2013bundled}\cite{liu2016meshflow}\cite{zhang2023minimum}. These traditional metrics try to estimate the spatial transformation (homography or affine) between adjacent frames from keypoint correspondences. However, the estimated point correspondences are unreliable in our challenging testing cases (\textit{e.g.}, low texture or low light). In addition, as described in Nie \textit{et\ al.}’s video stitching \cite{nie2017dynamic}, the metric in the frequency domain (\textit{i.e.}, the stability score in \cite{liu2013bundled}) are not reliable sometimes as the trajectory signals are usually very short and of different lengths. Therefore, we adopt the more intuitive indicators (\textit{i.e.}, the distortion loss and smoothness loss) to describe the distortion and stability performance. 

\begin{table}[!t]
	\begin{center}
		\renewcommand{\arraystretch}{1.1}
		\caption{More ablation studies about the optimization components of the warp smoothing model on alignment performance ($\uparrow$).}
        \vspace{-0.3cm}
		\label{alignment}
    \scalebox{0.8}{
		\begin{tabular}{ccccccc}
			\hline
			 &Method & Regular & Low-Light & Low-Texture & Moving-Fast & Average  \\
			\cline{2-7}
            1 & Only Spatial Warp  & 25.60/0.851 & 35.18/0.960 & 33.92/0.928& 25.57/0.840 & 30.75/0.903 \\
            2 & w/o Overlapping Mask ($OP$)  & 22.26/0.759 & 32.14/0.945 & 30.97/0.913& 20.73/0.713 & 27.40/0.851 \\
			3 & w/o Smoothness ($\mathcal{L}_{smothness}$)  & 25.61/0.851 & 35.18/0.960 & 33.92/0.928 & 25.57/0.840 & 30.75/0.903 \\
            4 &  w/o Spatial Consistency ($\mathcal{L}_{space}$)  & 24.64/0.832 & 34.49/0.958 & 33.62/0.927 & 23.39/0.788 & 29.89/0.890 \\
            5 & w/o Online Collaboration ($\mathcal{L}_{online}$)  & 24.70/0.833 & 34.50/0.958 & 33.62/0.927 & 23.39/0.787 & 29.91/0.890 \\
            6 & \textit{StabStitch}  & 24.64/0.832 & 34.51/0.958 & 33.63/0.927 & 23.36/0.787 & 29.89/0.890 \\
			\hline
		\end{tabular}
  }
	\end{center}
 \vspace{-0.5cm}
\end{table}

\begin{table}[!t]
	\begin{center}
		\renewcommand{\arraystretch}{1.1}
		\caption{More ablation studies about the optimization components of the warp smoothing model on distortion performance ($\downarrow$).}
        \vspace{-0.3cm}
		\label{distortion}
    \scalebox{0.86}{
		\begin{tabular}{ccccccc}
			\hline
			 &Method & Regular & Low-Light & Low-Texture & Moving-Fast & Average  \\
			\cline{2-7}
            1 & Only Spatial Warp  & 0.925 & 0.913 & 0.767 & 0.610 & 0.804 \\
            2 & w/o Overlapping Mask ($OP$)  & 0.624 & 0.566 & 0.439 & 0.509 & 0.535 \\
			3 & w/o Smoothness ($\mathcal{L}_{smothness}$)  & 0.554 & 0.598 & 0.471 & 0.514 & 0.534 \\
            4 &  w/o Spatial Consistency ($\mathcal{L}_{space}$)  & 1.189 & 1.214 & 1.081 & 1.098 & 1.145 \\
            5 & w/o Online Collaboration ($\mathcal{L}_{online}$)  & 0.682 & 0.695 & 0.591 & 0.796 & 0.691 \\
            6 & \textit{StabStitch}  & 0.661 & 0.660 & 0.638 & 0.735 & 0.674 \\
			\hline
		\end{tabular}
  }
	\end{center}
 \vspace{-0.5cm}
\end{table}

\begin{table}[!t]
	\begin{center}
		\renewcommand{\arraystretch}{1.1}
		\caption{More ablation studies about the optimization components of the warp smoothing model on stability performance ($\downarrow$).}
        \vspace{-0.3cm}
		\label{stability}
    \scalebox{0.845}{
		\begin{tabular}{ccccccc}
			\hline
			 &Method & Regular & Low-Light & Low-Texture & Moving-Fast & Average  \\
			\cline{2-7}
            1 & Only Spatial Warp  & 29.03 & 26.33 & 27.35 & 158.57 & 60.32 \\
            2 & w/o Overlapping Mask ($OP$)  & 22.22 & 18.06 & 15.68 & 129.00 & 46.24\\
			3 & w/o Smoothness ($\mathcal{L}_{smothness}$)  & 28.98 & 26.34 & 27.35 & 158.22 & 60.22 \\
            4 &  w/o Spatial Consistency ($\mathcal{L}_{space}$)  & 23.38 & 19.60 & 18.67 & 133.85 & 48.88 \\
            5 & w/o Online Collaboration ($\mathcal{L}_{online}$)  & 23.53 & 19.69 & 18.78 & 135.25 & 49.33 \\
            6 & \textit{StabStitch}  & 23.18 & 19.53 & 18.67 & 133.59 & 48.74 \\
			\hline
		\end{tabular}
  }
	\end{center}
 \vspace{-0.4cm}
\end{table}

\section{More Experiment}
\label{sec:6}

In this section, we conduct more experiments to explore the roles of different optimization components in the warp smoothing model.

We first report the performance of the spatial warp model and complete \textit{StabStitch}, and then ablate each constraint to show its effectiveness. The alignment, distortion, and stability performance are shown in Tab. \ref{alignment}, Tab. \ref{distortion}, and Tab. \ref{stability}, respectively.

\paragraph{Data Term:} 
The data term requires the smoothed trajectories to be close to the original trajectories. Without this term, the final output trajectories will degrade to constant paths, yielding meaningless results. Therefore, we ablate the overlapping mask ($OP$) instead by setting $\alpha$ to $0$. As depicted in Tab. \ref{alignment}, the alignment performance significantly decreases from 30.75/0903 to 27.40/0.851. In this case, extensive artifacts will be produced.

\paragraph{Smoothness Term:} 
The smoothness term works together with the data term to strike a balance between preserving the original trajectories (especially alignment performance) and smoothing the trajectories. Without this term, the output trajectories will be close to the original trajectories. As shown in Tab. \ref{alignment} and Tab. \ref{stability} (Experiment 1 and 3), the alignment and stability performance is close to that of the spatial warp model. As for the distortion performance reported in Tab. \ref{distortion}, it is significantly improved because of the spatial consistency term. If we further remove the spatial consistency term on the basis of Experiment 3, the distortion score will also approach that of the spatial warp model.

\paragraph{Spatial Consistency Term:} 
With only the data and smoothness terms, every trajectory will be optimized independently. However, there are $(U+1)\times (V+1)$ control points, which implies $(U+1)\times (V+1)$ trajectories. If each trajectory is smoothed separately without considering the consistency between trajectories, distortions are prone to occur. As shown in Tab. \ref{distortion} (Experiment 4 and 6), the distortion is significantly increased without this term.

\paragraph{Online collaboration Term:} 
In the online mode, only the last frame in a sliding window (containing $N$ frames) is used. The online collaboration term contributes to the stability of adjacent sliding windows. Without this term, the stability slightly degrades especially in the category of MF, as illustrated in Tab. \ref{stability} (Experiment 5 and 6). 

\vspace{0.2cm}
The final model (\textit{StabStitch}) does not achieve the best performance in alignment, distortion, and stability. But it reaches the best balance among the three metrics and produces the best visual effect, as demonstrated in our supplementary video.

\end{document}